\documentclass[12pt]{article}
\usepackage{graphicx}
\usepackage{amsmath}
\usepackage{hyperref}
\usepackage{geometry}
\usepackage{float}
\title{HackerRank-ASTRA: Evaluating Correctness \& Consistency of Large Language Models on Cross-Domain Multi-File Project Problems}
\author{Jun Xing, Mayur Bhatia, Sahil Phulwani, Darshan Suresh, Rafik Matta}
\date{}

\begin{document}

\maketitle

\begin{abstract}
Evaluating the real-world applicability of large language models (LLMs) provides valuable insights for their development and use in software development tasks. Existing benchmarks often focus on standalone coding problems or specific libraries, overlooking multi-file, project-based scenarios and lacking a rigorous evaluation for consistency. The HackerRank-ASTRA Benchmark introduces project-based coding problems that mirror real-world scenarios. It evaluates model consistency through 32 runs ($k=32$) and median standard deviation while incorporating taxonomy-level analysis to assess sub-skill capabilities. Initial evaluations (v1) on 65 problems show that the top three models---o1, o1-preview, and Claude-3.5-Sonnet-1022---achieved comparable average scores of 75\%, with no statistically significant differences in performance. Notably, Claude-3.5-Sonnet-1022 demonstrated the highest consistency across problems, with remarkably low variability (SD = 0.0497), which was statistically significant compared to other models, highlighting its reliability for real-world software development tasks.
\end{abstract}

\section{Introduction}
The rapid advancement of large language models (LLMs) has significantly impacted software development, enabling capabilities such as code generation and bug fixing. However, evaluating these models' real-world effectiveness is challenging. Many of today’s most popular and widely used evaluation benchmarks are heavily focused on a single language or single-file, well-defined tasks. For instance, HumanEval focuses on standalone coding tasks, evaluating models for generating single-function solutions without accounting for multi-file dependencies or broader project contexts\cite{yu2024humaneval}. SWE-bench introduces GitHub-based evaluations for resolving real-world issues but focuses on 12 specific Python libraries\cite{jimenez2023swebench} and 17 JavaScript Repositories \cite{yang2024swebench}. DevEval broadens the domain scope by introducing multi-file coding tasks that simulate the software development lifecycle, including software design and testing \cite{li2024deveval}. However, despite its breadth, DevEval does not explicitly evaluate model consistency across multiple runs, leaving a critical gap in understanding LLM reliability for real-world applications.

Moreover, a study by \cite{huang2024enhancing} introduces the concept of self-consistency in Code LLMs, emphasizing that a trustworthy model should generate consistent natural language specifications for the code it generates and vice versa. Their evaluation of eleven code LLMs reveals frequent failures in maintaining self-consistency, highlighting a gap between traditional accuracy metrics and the models' true understanding of the shared semantics between natural and programming languages. This underscores the need for more robust evaluation frameworks that go beyond conventional metrics to assess the reliability and predictability of LLM-generated code.

To address these limitations, the HackerRank-ASTRA (Assessment of Software Tasks in Real-world Applications) Benchmark offers a comprehensive evaluation framework for multi-file, project-based software development problems. ASTRA’s initial release (v1) focuses on frontend development, featuring frameworks such as Node.js, React.js, Angular.js, Django, Java Spring Boot, Ruby on Rails, and .NET. The benchmark evaluates new feature development, where both inputs and outputs are text-based. Metrics such as mean pass@1 and mean score assess model correctness, while median standard deviation across 32 runs ($k=32$) provides insights into consistency and reliability. By simulating practical coding challenges, HackerRank-ASTRA aims to provide actionable insights into the capabilities and limitations of state-of-the-art LLMs in addressing modern software engineering needs.

\section{HackerRank-ASTRA}
HackerRank-ASTRA is a benchmark built from HackerRank’s proprietary library of multi-file, project-based software development problems. These problems were originally designed to assess the software development skills of human developers across a wide range of skill domains in realistic, project-like settings. We observed that even advanced large language models (LLMs) face significant challenges when solving these problems, which motivated the creation of this benchmark.

\subsection{Task Formulation}
The task requires models to take both the problem statements and relevant source code files as input, with the objective of generating the requested code as output. To assess consistency, the process is repeated multiple times, with each run initialized as a new conversation to eliminate prior memory or contextual bias. Model performance is evaluated based on the percentage of test cases passed, ensuring a clear and objective scoring system. Since these problems were originally intended for human evaluation, they include a diverse and carefully curated set of test cases, with an average of 6.7 test cases per problem in the current version (v1). This content was preapred by HackerRank's Content Creation team.

\subsection{Features of HackerRank-ASTRA}
\textbf{Diverse Skill Domains:} The v1 ASTRA Benchmark dataset comprises 65 project-based coding problems, primarily focused on front-end development. These problems are categorized into 10 primary coding skill domains and 34 subcategories (Additional technical details are in Section \ref{skilldomains}), ensuring a comprehensive evaluation of diverse technical capabilities.

\textbf{Long Context Multi-File Project Questions:} To closely replicate real-world software development tasks, the HackerRank-ASTRA Benchmark dataset includes, on average, 12 source code and configuration files per question as model inputs. The problem statements alone have an average context length of 718 characters, while source code inputs average 22,863 characters, and output strings average 2,744 characters. On average, the benchmark requires 84 lines of solution code per problem and modifications to 2.3 files, reflecting the complexity and breadth of real-world project scenarios.

\textbf{Model Correctness and Consistency Evaluation:} To evaluate production reliability, the benchmark prioritizes metrics such as mean scores, mean pass@1, and median standard deviation with $k=32$, rather than relying solely on the traditional industry-standard pass@$k$. These metrics provide a more nuanced assessment of the model's correctness and consistency, crucial for real-world applicability.

\textbf{Wide Test Case Coverage:} The ASTRA Benchmark dataset includes an average of 6.7 test cases per question, designed to rigorously evaluate the correctness of model-generated solutions across a variety of implementation scenarios.

\textbf{Real-world Alignment:} HackerRank’s extensive content library serves as the foundation for the ASTRA Benchmark. With over 7,500 coding questions, an advanced skills taxonomy, and data-driven insights, HackerRank has well-established expertise in evaluating developer skills. Informed by data from tens of thousands of job descriptions, HackerRank’s Roles Directory spans 9 job families, 77 roles, and 260 skills. This ensures alignment with real-world industry demands, leveraging machine learning to identify key skills critical for various technical roles.

\subsection{Skill Domains \label{skilldomains}}
The v1 ASTRA Benchmark consists of 65 project-based coding questions, systematically organized into 10 primary skill domains and 34 sub-skill categories.
\begin{figure}[H]
    \centering
    \includegraphics[width=0.8\textwidth]{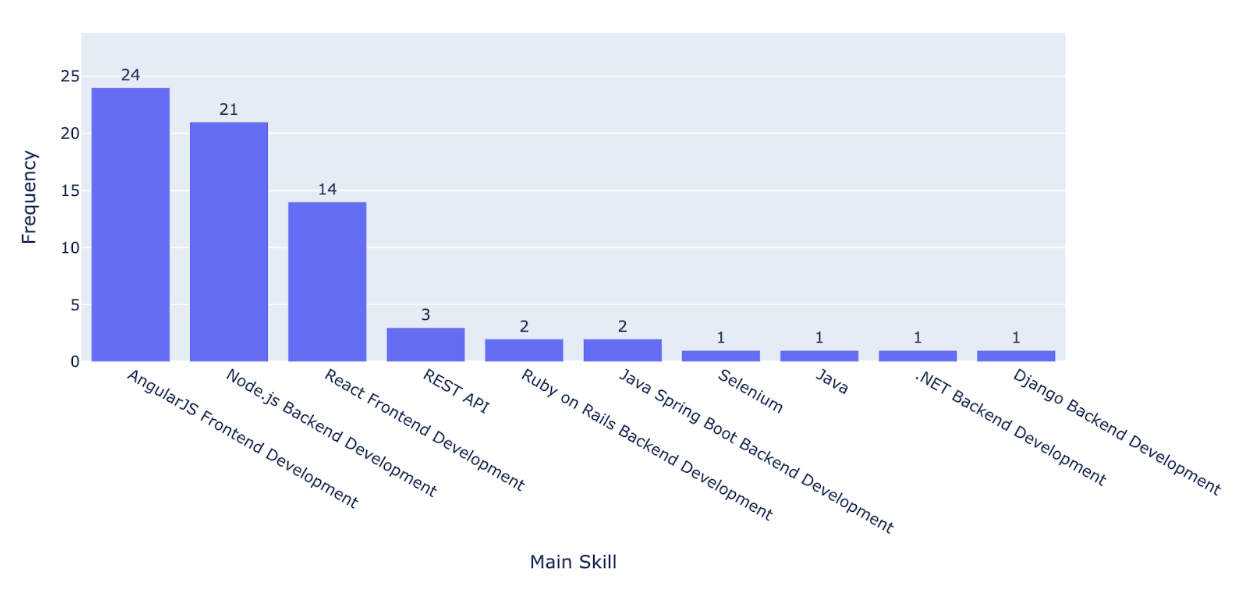} 
    \caption{Distribution of v1 HackerRank-ASTRA benchmark main skill frequency.}
    \label{fig:placeholder1}
\end{figure}

\begin{figure}[H]
    \centering
    \includegraphics[width=0.8\textwidth]{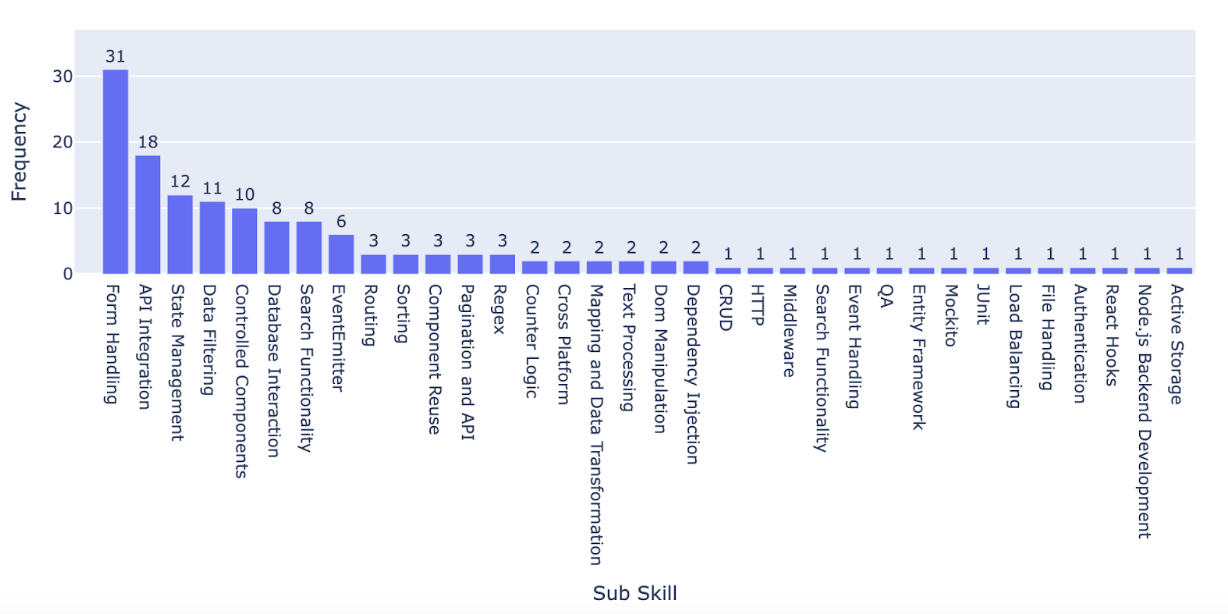} 
    \caption{Distribution of v1 HackerRank-ASTRA benchmark sub-skill frequency.}
    \label{fig:placeholder2}
\end{figure}

\subsection{Key Statistics}

\begin{table}[H]
    \centering
    \caption{Key Statistics of v1 HackerRank-ASTRA Benchmark}
    \begin{tabular}{|l|c|}
        \hline
        \textbf{Statistic} & \textbf{Value} \\
        \hline
        Total Project Questions & 65 \\
        Count of Main Skill Categories & 10 \\
        Count of Sub Skill Categories & 34 \\
        Average Number of Test Cases & 6.7 \\
        Average Input Files & 12 \\
        Average Input Character Length & 22,863 \\
        Average Problem Statement Character Length & 718 \\
        Average Output Character Length & 2,744 \\
        Average Expected Lines of Code & 84 \\
        Average Modified Code Files & 2.3 \\
        \hline
    \end{tabular}
    \label{tab:key_statistics}
\end{table}

\subsection{Evaluation Metrics}
\textbf{Mean Score:} The Mean Score ($\frac{\text{Passed Test Cases}}{\text{Total Test Cases}}$) with $k=32$ evaluates the model’s partial correctness and robustness by considering multiple attempts (up to $k=32$) for each problem. For each problem, the score is calculated as the average proportion of passed test cases across the 32 runs. Then, this score is aggregated across 65 problems to compute the final Mean Score. Formally:
\begin{equation}
    \text{Mean Score} = \frac{1}{n} \sum_{i=1}^{n} \frac{1}{k} \sum_{j=1}^{k} \frac{p_{ij}}{T_i},
\end{equation}

\textbf{Mean Pass@1:} The Mean pass@1 metric evaluates the frequency with which the model achieves a perfect score across $k=32$ runs for each problem. For a given problem, pass@1 is defined as the proportion of runs where the model achieves a perfect score (all test cases passed). The metric then aggregates this proportion across $n=65$ problems to compute the final Mean pass@1:
\begin{equation}
    \text{Mean Pass@1} = \frac{1}{n} \sum_{i=1}^{n} \frac{1}{k} \sum_{j=1}^{k} I\left(\frac{p_{ij}}{T_i} = 1\right),
\end{equation}

\textbf{Consistency:} Consistency is defined as the median standard deviation of the scores of the model’s output across $n=65$ problems. For each problem, the standard deviation of scores across $k=32$ solutions is computed. The final metric is the median of these standard deviations across all problems. This approach is used because the standard deviation of scores across different problems often deviates from a normal distribution. A lower median standard deviation indicates more consistent performance, while a higher median standard deviation suggests greater variability.
\begin{enumerate}
    \item Compute the standard deviation of scores for each problem $i$:
    \begin{equation}
        SD_i = \sqrt{\frac{1}{k} \sum_{j=1}^{k} \left( \text{Score}_{ij} - \overline{\text{Score}_i} \right)^2}
    \end{equation}
    Where 
    \begin{equation}
        \overline{\text{Score}_i} = \frac{1}{k} \sum_{j=1}^{k} \text{Score}_{ij}
    \end{equation}
    is the mean score for problem $i$.
    
    \item Compute the Median SD across $n = 65$ problems:
    \begin{equation}
        \text{Median SD} = \text{median} \left(SD_1, SD_2, \dots, SD_n \right)
    \end{equation}
\end{enumerate}

These metrics are chosen for their alignment with real-world software development standards, where both complete and partially correct solutions carry significance. The Mean Score accounts for the model’s incremental problem-solving ability, offering a granular view of how much of a solution’s functionality is achieved even when it is not fully correct. Pass@1 indicates how reliably a model can produce correct code in a single shot, which is crucial in real-world scenarios where developers working with code LLMs want to apply minimal revisions to the output. The consistency of a model’s solutions for each problem highlights whether the model performs steadily across its multiple attempts or exhibits significant variability.
Using k=32 provides a meaningful measure of a model’s capability to explore diverse solutions, as this number of attempts allows it to overcome minor variances while maintaining focus on a feasible solution space.
We use the mean for metrics like mean score and pass@1 because these aggregate metrics aim to capture the overall performance of the model across problems. For standard deviation, however, we use the median because the variability of scores across problems often contains outliers, and the median provides a more robust measure of the typical consistency of the model's performance.

\subsection{Sample Problem}
\begin{figure}[H]
    \centering
    \includegraphics[width=0.8\textwidth]{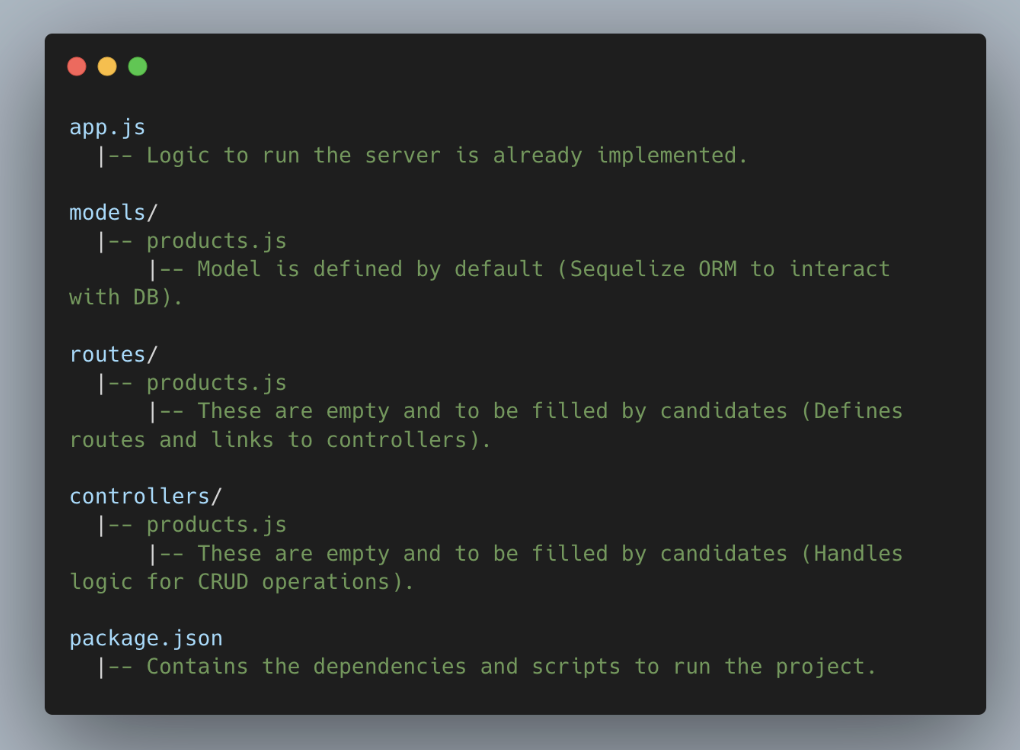} 
    \caption{Project structure of a sample RESTful API problem.}
    \label{fig:sample_project}
\end{figure}

The following is an example of a RESTful API project from the ASTRA Benchmark Dataset. This task involves developing a RESTful API for managing product records using Node.js and Express, simulating a real-world e-commerce development scenario. The project structure is depicted in Figure \ref{fig:sample_project}.

This problem prepares models for practical API development challenges, emphasizing correctness, maintainability, and scalability. Additional technical details are included in the Appendix.

\section{Experiment Setup}
\subsection{Evaluation Criteria}
The evaluation primarily targets code generation correctness and consistency, focusing exclusively on the model’s ability to generate accurate and functional solutions in response to a text-based API call.

\subsection{Evaluation Pipeline}
\begin{figure}[H]
    \centering
    \includegraphics[width=0.8\textwidth]{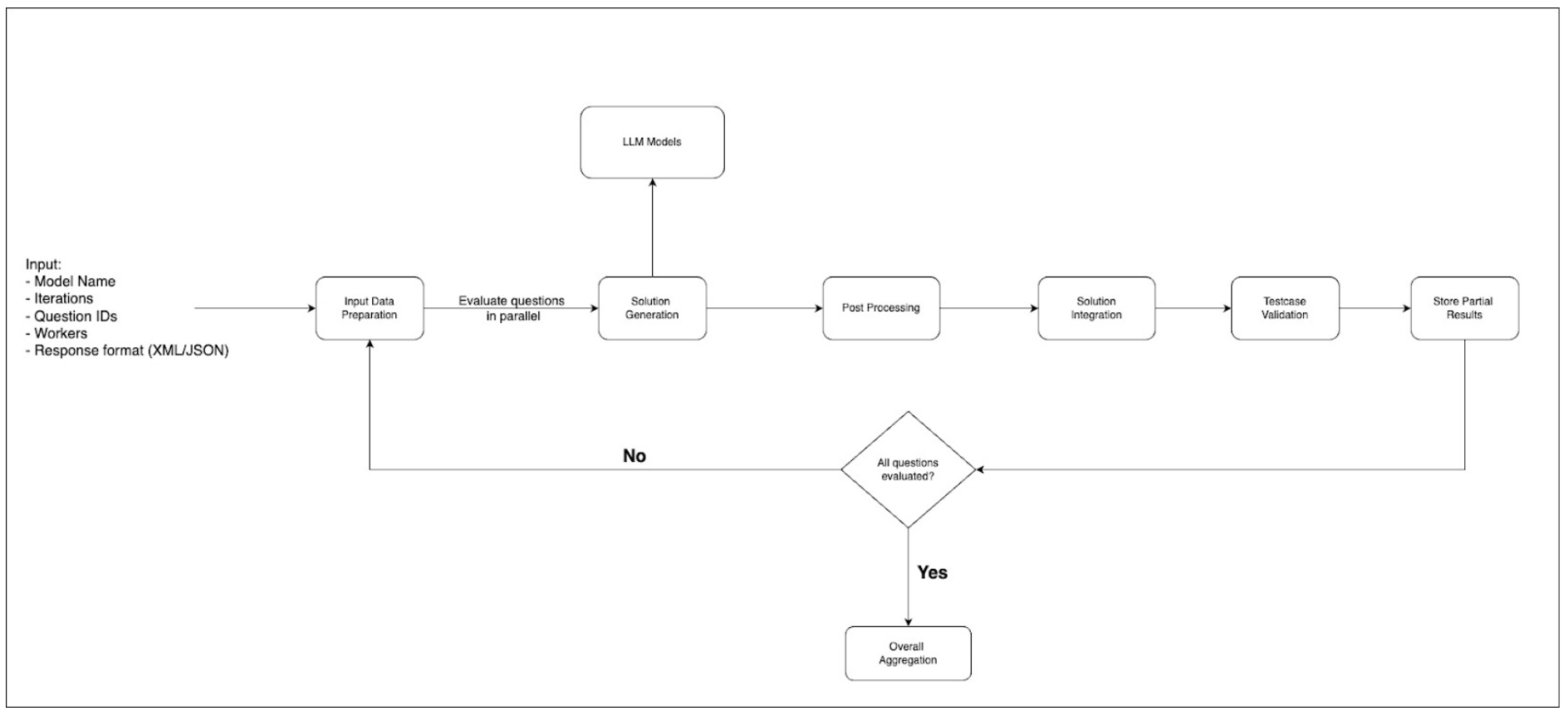} 
    \caption{Diagram of v1 HackerRank-ASTRA benchmark evaluation pipeline.}
    \label{fig:pipeline_diagram}
\end{figure}

\textbf{Input Data Preparation:} The evaluation process begins by reading a CSV file containing the list of questions. For each question, the corresponding project files are retrieved from an S3 bucket, ensuring that input data is complete and consistent. This step eliminates ambiguity and standardizes the input for the model. A structured prompt is then created, comprising:
\begin{itemize}
    \item \textbf{Question Instructions:} Concise directives for the task.
    \item \textbf{Problem Statement:} A detailed description of the coding challenge.
    \item \textbf{Project Files:} Relevant source files necessary for the solution.
\end{itemize}

\textbf{Solution Generation:} The structured prompt is sent to the selected AI model(s) (e.g., o1, Gemini, Claude). The model generates a solution, which is returned in the specified format (e.g., XML or JSON).

\textbf{Post-Processing:} The generated solution is validated for structural and formatting issues, such as:
\begin{itemize}
    \item Parsing errors in XML/JSON.
    \item Misformatted line breaks (\textbackslash n) or tabs (\textbackslash t).
    \end{itemize}
Necessary corrections are applied to ensure the solution adheres to the required structure and remains parsable.

\textbf{Solution Integration:} The validated solution is integrated into the project files, updating the project with the model’s output and preparing it for testing.

\textbf{Test Case Validation:} The updated project is executed in a Docker container, where it is evaluated against pre-defined test cases. These test cases serve as the ground truth to assess the correctness and consistency of the solution.

\textbf{Store Partial Results:} Evaluation results for each question, including the number of test cases passed and corresponding outputs, are recorded in a CSV file for further analysis.

\textbf{Overall Aggregation:} After evaluating all questions, an aggregation script computes key performance metrics for each question, including:
\begin{itemize}
    \item Mean score
    \item Mean Pass@1
    \item Standard deviation
    \item Mean test cases passed (used to calculate mean scores)
\end{itemize}

\subsection{Prompt}
In our prompt, we instruct the models to generate output code in either XML or JSON format. The primary motivation for adopting XML/JSON is that the HackerRank-ASTRA benchmark requires the model to handle multi-file modifications. Utilizing these structured formats ensures that the output is consolidated into a single file, facilitating consistent and reliable evaluation of the results.

As illustrated in the following figures, the JSON prompt included more detailed formatting instructions, as we observed format escaping issues with models generating JSON outputs. Since the objective is to evaluate the models' code generation capabilities rather than their proficiency in consolidating XML/JSON formats, we incorporated a post-processing step in the evaluation pipeline. This step standardizes the output formatting to ensure no bias is introduced due to formatting discrepancies.

\begin{figure}[H]
    \centering
    \includegraphics[width=0.8\textwidth]{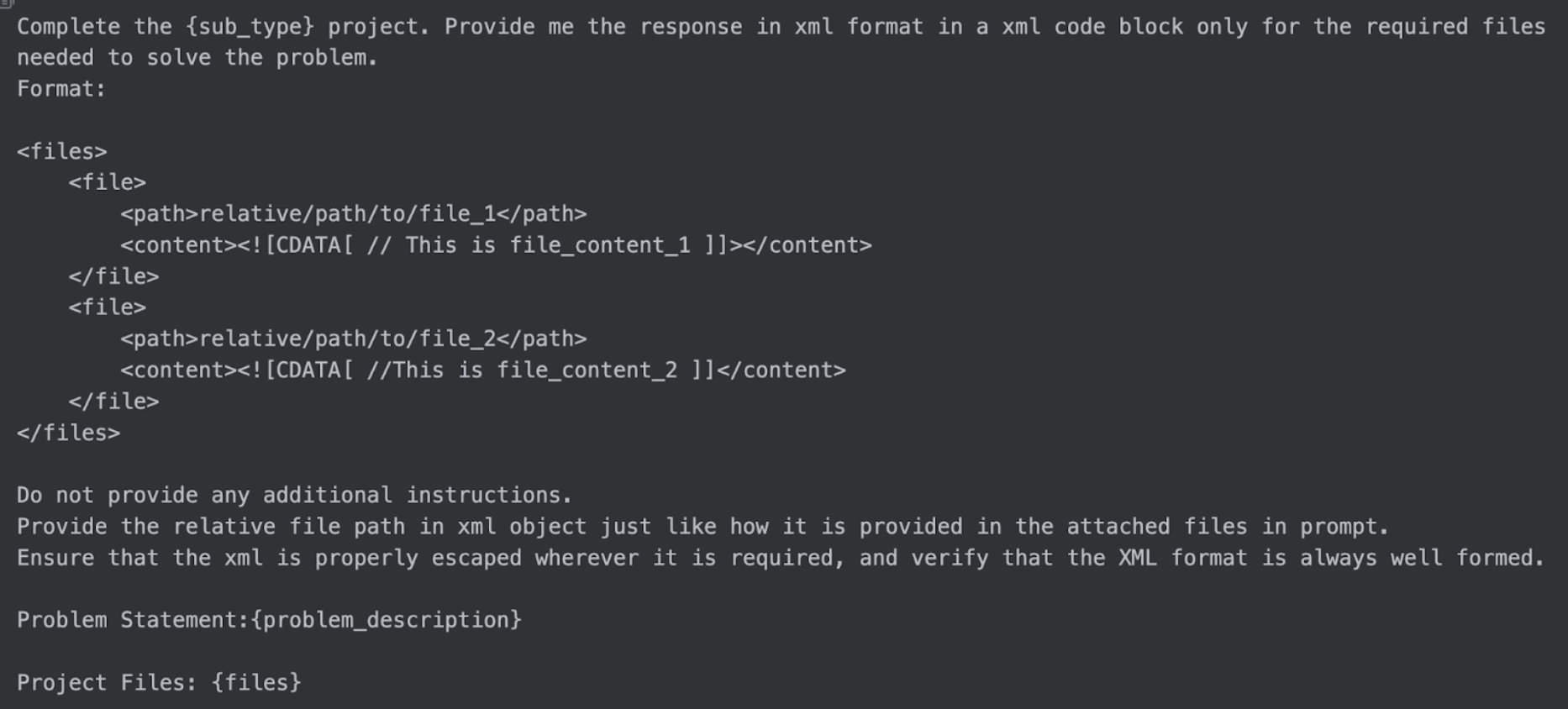} 
    \caption{XML prompt of v1 HackerRank-ASTRA benchmark.}
    \label{fig:xml_prompt}
\end{figure}

\begin{figure}[H]
    \centering
    \includegraphics[width=0.8\textwidth]{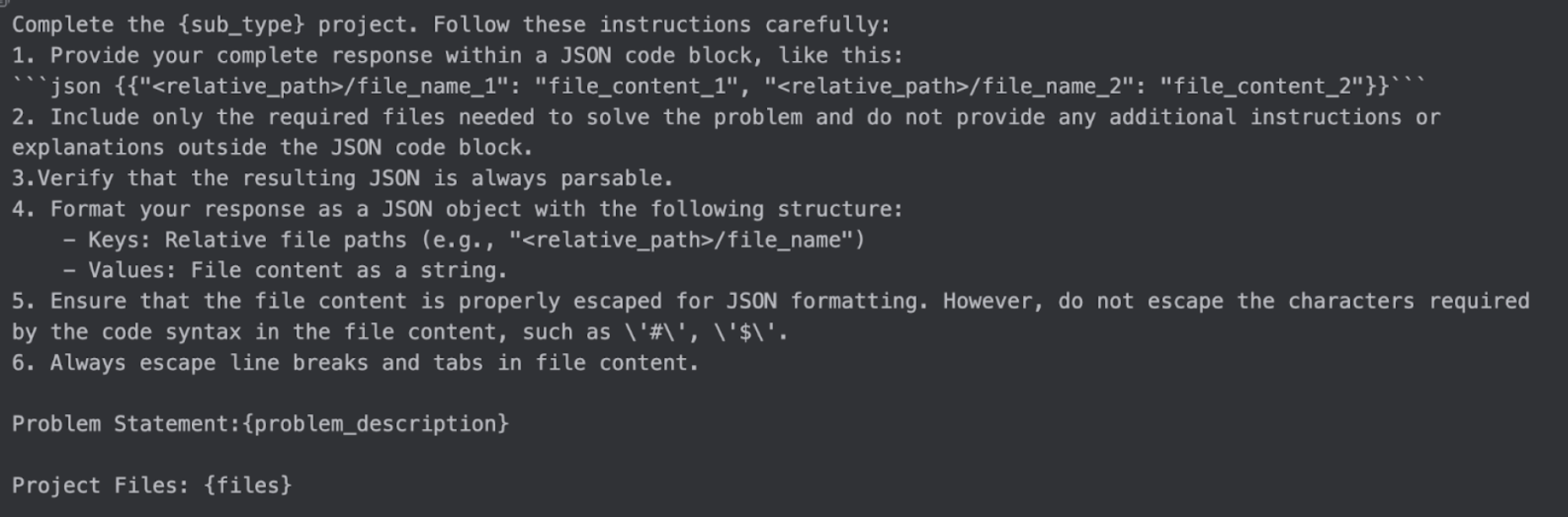} 
    \caption{JSON prompt of v1 HackerRank-ASTRA benchmark.}
    \label{fig:json_prompt}
\end{figure}
\subsection{Models}
In the v1 HackerRank ASTRA benchmark, we focused on evaluating frontier models, as detailed in the following tables. Given that our benchmark is designed to assess performance on real-world multi-file tasks, we have also emphasized the context length capabilities of the models to highlight their ability to handle complex scenarios effectively.

\begin{table}[H]
    \centering
    \caption{Context length and default temperature of models.}
    \begin{tabular}{|l|c|c|}
        \hline
        \textbf{Models} & \textbf{Context Length} & \textbf{Default Temperature} \\
        \hline
        o1 & 200,000 tokens & 1 \\
        o1-preview & 128,000 tokens & 1 \\
        GPT-4o-0513 & 128,000 tokens & 1 \\
        Claude-3.5-Sonnet-1022 & 200,000 tokens & 1 \\
        Gemini-1.5-pro & 128,000 tokens & 1 \\
        \hline
    \end{tabular}
    \label{tab:model_context_length}
\end{table}

\section{Results}

\subsection{Evaluation Leaderboard}
V1 ASTRA evaluates models on real multi-file front-end challenges, yielding an mean score around 70\% and an mean pass@1 around 60\%. Based on the analysis of mean scores, the models o1, o1-preview, and Claude-3.5-Sonnet-1022 demonstrate superior performance for multi-file, real-world front-end tasks. However, due to the high variance within the mean scores across 65 questions, a paired t-test reveals that, with the exception of GPT-4o-0513, the differences between model performances are not statistically significant. Despite this, the mean score with $k=32$ indicates a meaningful practical impact in real-world production settings. Similar trends were observed when evaluating the models using the mean pass@1 metric.

In our benchmark evaluation, we assessed the consistency of LLMs using the standard deviation (SD) of their scores across 32 independent runs per question and then evaluated the median SD across 65 questions. The models demonstrated varying levels of performance stability, with Claude-3.5-Sonnet-1022 exhibiting the lowest variability (SD = 0.0497), indicating the highest consistency across problems. The difference between Claude-3.5-Sonnet-1022 and the rest of the models is statistically significant based on the paired t-test.

\begin{table}[H]
    \centering
    \caption{Model performance comparison for the v1 HackerRank-ASTRA benchmark.}
    \begin{tabular}{|l|c|c|c|}
        \hline
        \textbf{Model} & \textbf{Mean Score} & \textbf{Consistency (SD)} & \textbf{Mean Pass@1} \\
        \hline
        o1 & 75.80\% & 0.11 & 63.92\% \\
        o1-preview & 75.55\% & 0.17 & 60.89\% \\
        Claude-3.5-Sonnet-1022 & 75.07\% & 0.05 & 62.74\% \\
        Gemini-1.5-pro & 71.17\% & 0.13 & 58.15\% \\
        GPT-4o-0513 & 69.52\% & 0.20 & 58.15\% \\
        \hline
    \end{tabular}
    \label{tab:model_performance}
\end{table}

\subsection{Main Skill Model Performance Summary}
The best model for front-end development depends on the specific skills being evaluated. For skills with an occurrence of $\geq3$ in the benchmark, o1, o1-preview, and Claude-3.5 Sonnet demonstrated comparable performance levels. However, for skills with only a single occurrence in the ASTRA dataset, such as Java and Selenium, Claude-3.5 Sonnet and Gemini 1.5 Pro tended to outperform OpenAI models.

Interestingly, o1 underperformed compared to its predecessor, o1-preview, and even GPT-4o for certain skills, including AngularJS, Java Spring Boot Backend, Java, and Selenium. Notably, o1-preview achieved the highest performance across all models on tasks involving AngularJS, proving that newer iterations do not always guarantee better results.

\begin{figure}[H]
    \centering
    \includegraphics[width=0.8\textwidth]{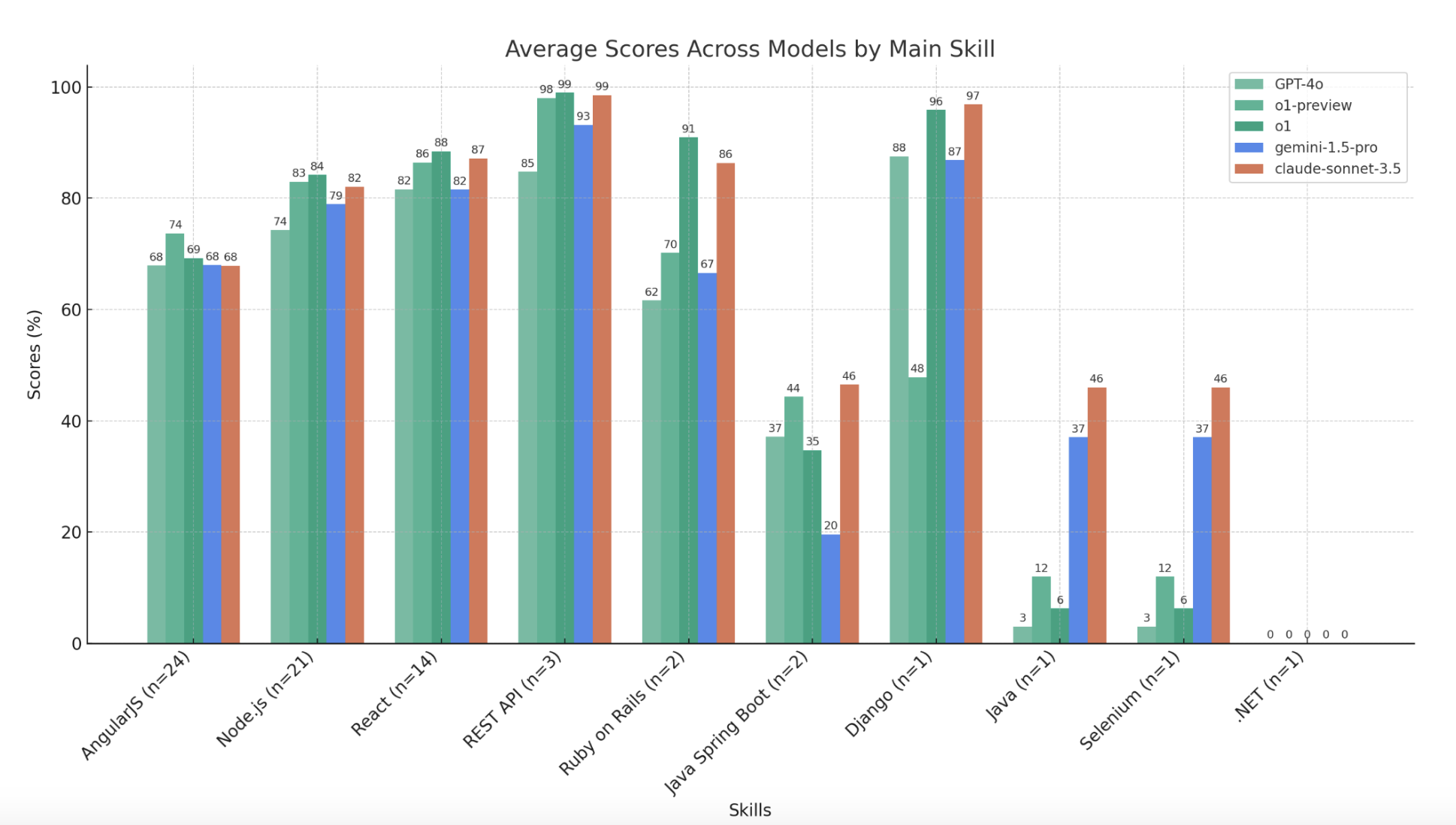} 
    \caption{Model performance comparison by main skill categories.}
    \label{fig:main_skill_performance}
\end{figure}

\subsection{Sub-Skill Model Performance Summary}
Model performance exhibits significant variation across subskills, indicating that the optimal AI tool for front-end development is highly dependent on the specific use case. Claude 3.5 Sonnet demonstrates superior performance in areas such as API integration, data filtering, and database interaction. Conversely, o1 performs particularly well in tasks related to form handling, pagination, API management, and EventEmitter functionality. Notably, o1-preview and GPT-4o outperform o1 in several subskills, underscoring the observation that newer models do not consistently achieve superior performance across all domains. These findings highlight the necessity of selecting models based on the precise requirements of individual projects to achieve optimal outcomes.

\begin{table}[H]
    \centering
    \caption{Winning models by sub-skill categories.}
    \begin{tabular}{|l|l|}
        \hline
        \textbf{Subskill (Occurrence)} & \textbf{Winning Models} \\
        \hline
        Form Handling (31) & o1 \\
        API Integration (18) & Claude-3.5-Sonnet, o1 \\
        State Management (12) & Claude-3.5-Sonnet \\
        Data Filtering (11) & Claude-3.5-Sonnet \\
        Controlled Components (10) & Gemini-1.5-Pro \\
        Search Functionality (9) & o1-preview \\
        Database Interaction (8) & Claude-3.5-Sonnet \\
        EventEmitter (6) & o1 \\
        Component Reuse (3) & Claude-3.5-Sonnet \\
        Pagination and API (3) & o1 \\
        Regex (3) & o1-preview \\
        Routing (3) & GPT-4o \\
        Sorting (3) & Claude-3.5-Sonnet \\
        \hline
    \end{tabular}
    \label{tab:subskill_performance}
\end{table}

\begin{figure}[H]
    \centering
    \includegraphics[width=0.8\textwidth]{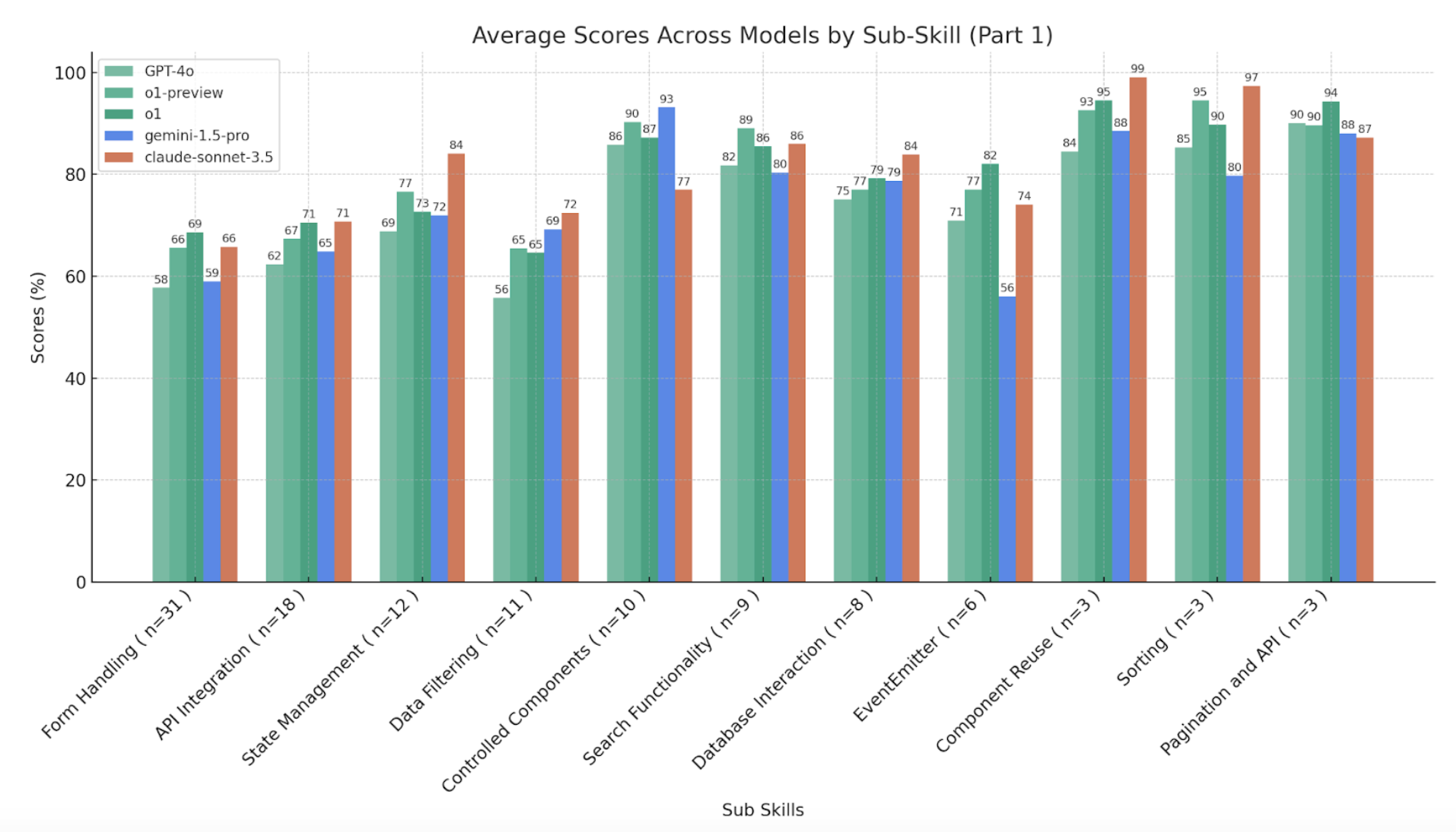} 
    \caption{Model performance comparison by sub-skill categories (1/3).}
    \label{fig:subskill_performance_1}
\end{figure}

\begin{figure}[H]
    \centering
    \includegraphics[width=0.8\textwidth]{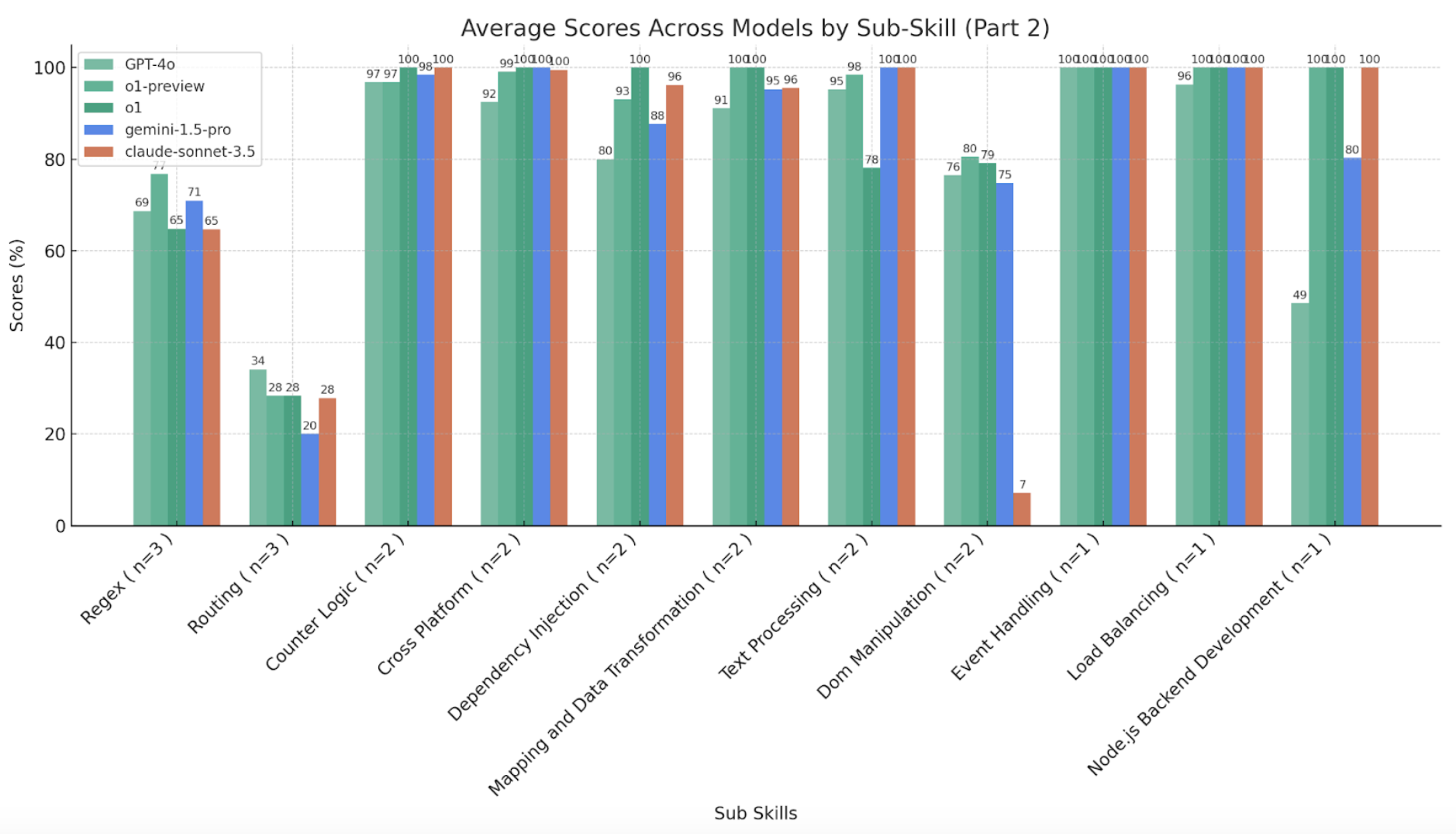} 
    \caption{Model performance comparison by sub-skill categories (2/3).}
    \label{fig:subskill_performance_2}
\end{figure}

\begin{figure}[H]
    \centering
    \includegraphics[width=0.8\textwidth]{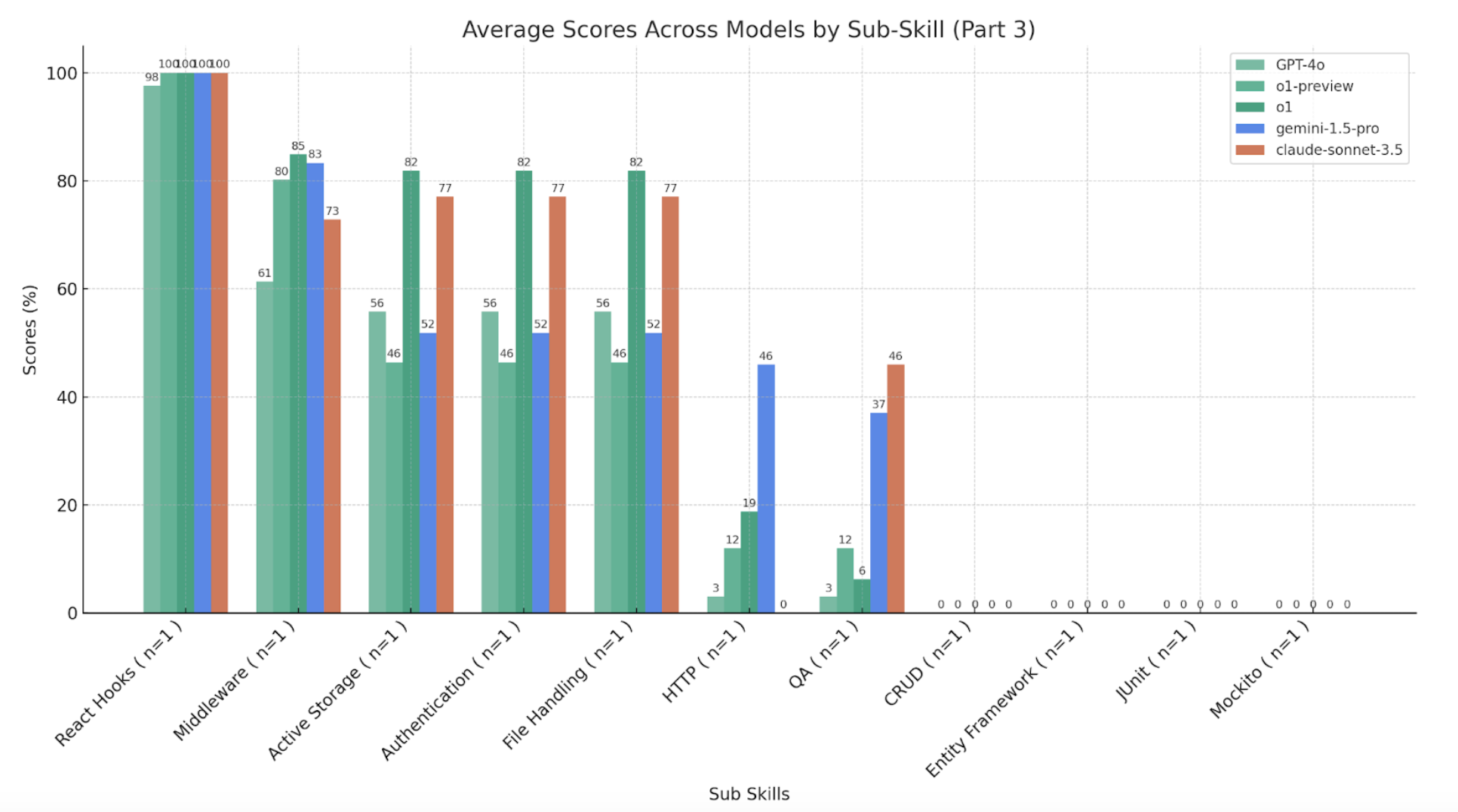} 
    \caption{Model performance comparison by sub-skill categories (3/3).}
    \label{fig:subskill_performance_3}
\end{figure}
\subsection{Formatting Impact on Model Performance}
XML consistently outperforms JSON across all evaluated models in our benchmark, which requires models to return multi-file code solutions in both XML and JSON formats. After assessing the mean score and mean pass@1 with $k=32$, XML demonstrated statistically significant superiority over JSON for both metrics, with the exceptions of GPT-4o and the mean score from Gemini 1.5 Pro. This suggests that there is likely more XML training data available for these large language models (LLMs) compared to JSON. For similar development scenarios, developers are advised to prioritize XML over JSON to achieve better results when leveraging LLMs.

\begin{table}[H]
    \centering
    \caption{XML format model results.}
    \begin{tabular}{|l|c|c|}
        \hline
        \textbf{Model} & \textbf{Mean Score} & \textbf{Mean Pass@1} \\
        \hline
        o1-preview & 75.56\% & 60.89\% \\
        Claude-3.5-Sonnet-1022 & 75.07\% & 62.74\% \\
        Gemini-1.5-Pro & 71.17\% & 58.15\% \\
        GPT-4o-0513 & 69.53\% & 50.91\% \\
        \hline
    \end{tabular}
    \label{tab:xml_results}
\end{table}

\begin{table}[H]
    \centering
    \caption{JSON format model results.}
    \begin{tabular}{|l|c|c|}
        \hline
        \textbf{Model} & \textbf{Mean Score} & \textbf{Mean Pass@1} \\
        \hline
        o1-preview & 72.36\% & 54.75\% \\
        Gemini-1.5-Pro & 70.08\% & 53.56\% \\
        Claude-3.5-Sonnet-1022 & 70.04\% & 57.50\% \\
        GPT-4o-0513 & 68.13\% & 50.31\% \\
        \hline
    \end{tabular}
    \label{tab:json_results}
\end{table}

\subsection{Format Escaping and Guardrail Issues}
The ASTRA Benchmark highlights challenges related to JSON escaping and occasional solution refusals in o1-preview and o1, underscoring the need for improved format escaping capabilities and guardrails. ASTRA includes multi-file project questions that require models to convert their outputs into JSON format. While most models handled this task seamlessly, o1-preview encountered notable difficulties, particularly with escaping multi-line strings. On average, o1-preview exhibited a 2.3\% error rate related to JSON escaping, even after detailed guidance and prompt adjustments. Additionally, o1-preview refused to provide solutions in 0.2\% of cases, likely due to the guardrail settings implemented within the model. Similar issues were observed with o1. These findings emphasize the need for refining guardrails to strike an optimal balance between security constraints and usability, ensuring more robust and reliable model performance in development tasks.

\subsection{Common Errors}
\begin{itemize}
    \item \textbf{User Interface and Presentation Issues:} Errors that impact the visual or interactive aspects of the application, degrading the user experience by displaying incorrect or suboptimal layouts and requiring user intervention to correct.
    \item \textbf{Logical and Implementation Errors:} Errors in the implementation that fail to account for specific conditions, edge cases, or problem constraints, despite having correct syntax.
    \item \textbf{Data Handling and Misuse Errors:} Errors caused by improper or unnecessary manipulation of data files or structures, disrupting the application's expected functionality and potentially leading to compilation failures.
    \item \textbf{Typos, Syntax, and Misinterpretation Errors:} Errors resulting from minor formatting issues, typographical mistakes, or misinterpretation of the problem statement. These errors typically involve incorrect output formatting or failure to adhere to the specified requirements.
\end{itemize}

\subsection{Correlation Between Model Performance and Input/Output Length}
The correlation between the average output length and the mean score is approximately -0.560, indicating a moderate negative relationship. This suggests that longer outputs are generally associated with lower scores. In contrast, the correlation between input length and mean score is approximately -0.164, reflecting a weak negative relationship. This implies that longer inputs may slightly reduce the mean score.

\begin{figure}[H]
    \centering
    \includegraphics[width=0.8\textwidth]{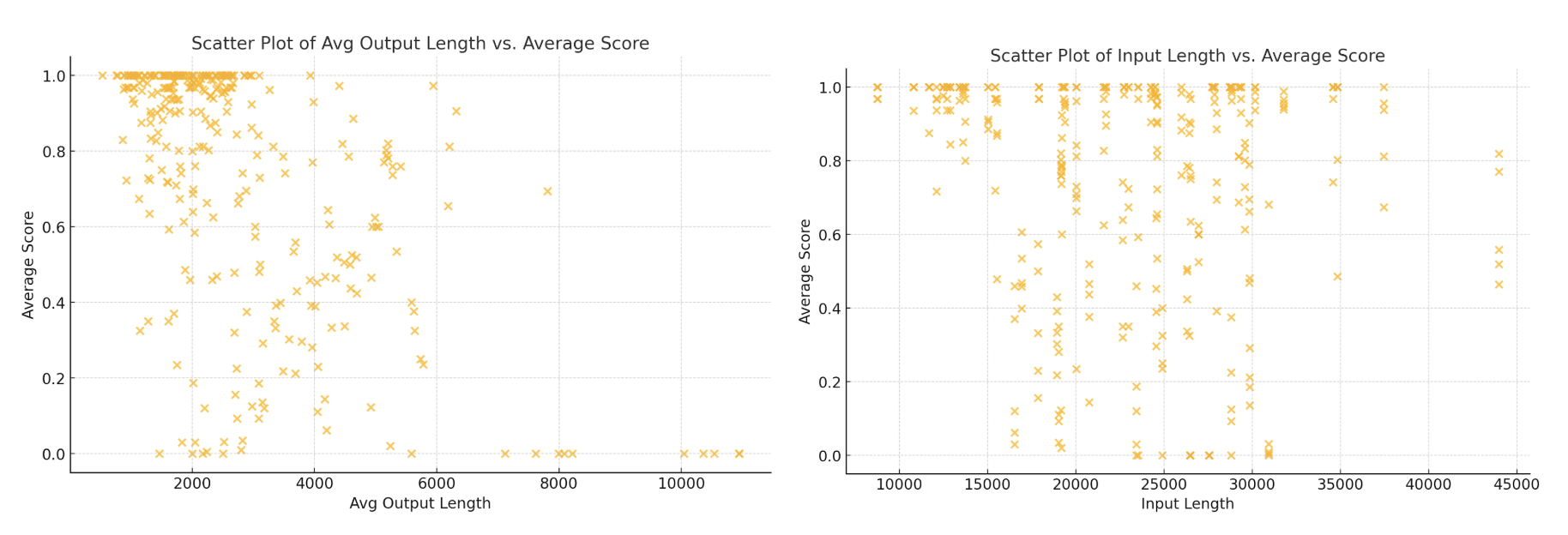} 
    \caption{Scatter Plot of Average Output vs Average Score vs Input Length.}
    \label{fig:correlation_plot}
\end{figure}

\section{Related Work}
The evaluation of large language models (LLMs) in coding tasks has evolved with the development of various benchmarks, each addressing specific aspects of model performance. Among the current industry standards, SWE-bench \cite{jimenez2023swebench} is notable for leveraging real-world Python repositories to assess model capabilities. Recently, SWE-bench expanded to include image data for multi-modal tasks, particularly targeting JavaScript challenges \cite{yang2024swebench}. DeepSeek-R1 \cite{deepseekai2024deepseekr1}sampled 16 responses for each question and calculated the overall average accuracy to ensure a stable evaluation, which essentially reflects the mean pass@1.

In recent years, repository-style benchmarks have emerged to cover a wider range of domains. For instance, DevEval \cite{li2024deveval}) evaluates LLMs on diverse repositories spanning various programming languages and domains, providing a more holistic assessment of coding capabilities.

The rise of agent-based benchmarks has further advanced the evaluation landscape. AgentBench \cite{liu2023agentbench} assesses LLMs' reasoning and decision-making abilities in interactive environments, offering insights into their performance as autonomous agents. Similarly, MLAgentBench\cite{huang2024mlagentbench} evaluates LLM-based research agents in machine learning tasks, focusing on their ability to perform experimentation loops and build models autonomously.

Our HackerRank-ASTRA Benchmark aligns with the trend of real-world evaluations by introducing a multi-file setup that reflects the complexity of production-level software development. Unlike most existing benchmarks, ASTRA emphasizes both correctness and consistency, evaluated through metrics like mean pass@1 and median standard deviation across multiple runs. Furthermore, ASTRA uniquely spans the full software development lifecycle (SDLC).

\section{Conclusions}
The HackerRank-ASTRA Benchmark contributes to the evaluation of LLMs in real-world software development by introducing a multi-file, project-based setup spanning a wide range of domains. By focusing on both correctness and consistency through metrics like mean pass@1 and median standard deviation, ASTRA offers deeper insights into model reliability and performance across repeated runs.

Initial findings reveal that LLMs face significant challenges in certain primary skills and sub-skill categories. Observations related to model consistency, formatting issues, guardrail adherence, and common error patterns further highlight areas where LLMs need improvement to better handle realistic and complex development scenarios.

\section{Limitations and Future Work}
While our study provides valuable insights into AI model performance on multi-file, real-world coding tasks, it has several limitations that warrant consideration. The current version of the benchmark focuses primarily on front-end projects, such as React and Angular.js, which narrows the scope of skill evaluation by under-representing back-end skills and other domains. Future iterations will address this by including a broader range of back-end technologies and skills. Additionally, the evaluation does not yet leverage argentic approaches, where models are granted autonomy to iteratively explore, adapt, and refine their solutions. Incorporating such methods would provide a more realistic assessment of model potential in dynamic problem-solving scenarios. Furthermore, the current framework evaluates models based on direct output without iterative feedback from test case results, limiting our ability to assess how models improve when guided by incremental corrections. Another limitation lies in the restricted model selection, which currently includes a subset of top-tier models. In future iterations, we plan to expand testing to include additional models, such as DeepSeek and Meta’s Llama models, and adopt a community-driven approach to benchmarking, fostering broader model comparisons. As a step toward this goal, we are open-sourcing all 65 project questions (https://huggingface.co/datasets/hackerrank/astra-benchmark) on GitHub and HuggingFace to enable wider participation and collaboration in advancing the benchmark.

\section{Acknowledgments}
We thank Vivek Ravisankar, the CEO and Co-Founder of HackerRank for his sponsorship of this work and valuable input. We acknowledge the insightful methodology discussions and support provided by Ahmed El-Kishky, Nat McAleese, and Chris Orsinger from OpenAI.

\bibliographystyle{unsrt}

\appendix
\renewcommand\thesection{A}
\renewcommand\thesubsection{A.\arabic{subsection}}

\section*{A Appendix}
\addcontentsline{toc}{section}{A Appendix}

\subsection{Errors Observed in a Restful project problem}
This example highlights a RESTful API development task from the ASTRA Benchmark Dataset, illustrating both the design requirements and the challenges encountered by language models like GPT-4. The task involves implementing a RESTful API for managing product records using Node.js and Express, mirroring real-world e-commerce application scenarios. Key endpoints include adding, retrieving, and updating product details, with explicit constraints against modifications or deletions using \texttt{PUT} or \texttt{DELETE} methods, reflecting specific business logic requirements. Additionally, candidates are required to ensure modular programming, robust error handling, and adherence to rules such as ensuring \texttt{MRP} $\geq$ \texttt{Price} and \texttt{Stock} $>$ 0 before publishing products.

A representative solution generated by GPT-4o-0513 demonstrates both the model's strengths and limitations. While it correctly implemented the core logic of the API by defining modular routes and controllers for handling product operations, a critical oversight occurred in integrating the routes into the main application file (\texttt{app.js}). Instead of linking the \texttt{productsRouter} to the root path, the model incorrectly retained a placeholder \texttt{indexRouter}, resulting in all requests to \texttt{/products} failing with a \texttt{404 Not Found} error. This error highlights a common gap in LLM outputs: an inability to ensure end-to-end functionality by resolving dependencies across modular components. Consequently, the model failed all test cases requiring functional routes.
\begin{figure}[H]
    \centering
    \includegraphics[width=0.8\textwidth]{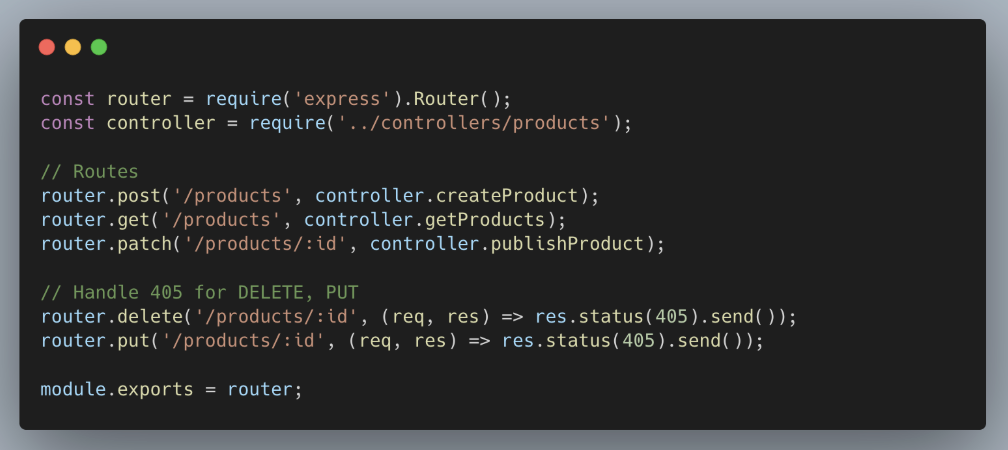} 
    \caption{The routes defined by GPT-4o-0513 for handling products.}
\end{figure}
\begin{figure}[H]
    \centering
    \includegraphics[width=0.8\textwidth]{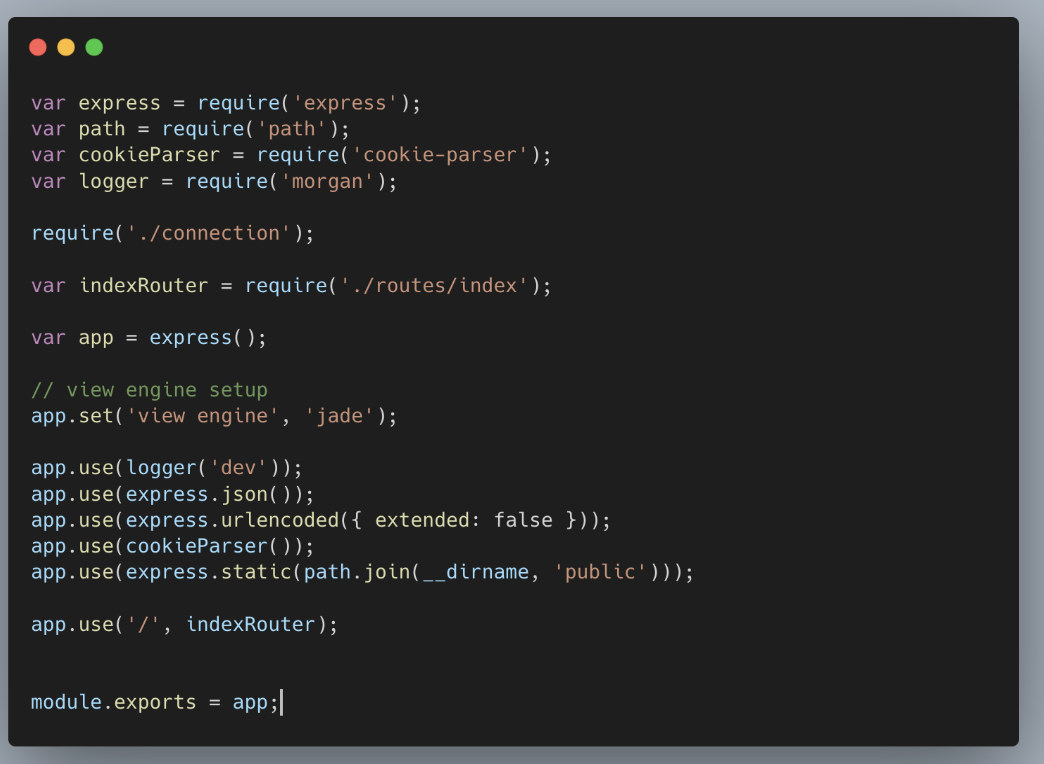} 
    \caption{The app.js provided by GPT-4o-0513.}
\end{figure}
\begin{figure}[H]
    \centering
    \includegraphics[width=0.8\textwidth]{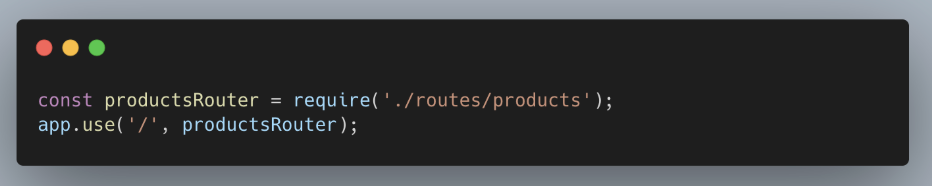} 
    \caption{Fixes required in app.js.}
\end{figure}
This example underscores the importance of validating LLM outputs in complex programming tasks. While models can generate syntactically correct code for modular components, they often miss essential integrations, leading to functionality gaps. Such errors reflect real-world challenges in software development, where end-to-end testing and integration are critical for delivering robust applications.

\renewcommand\thesubsection{A.\arabic{subsection}}

\subsection{Errors Observed in a React project problem}
This problem revolves around creating a dynamic form with two fields: "Company Title" and "Number of Employees." The task requires implementing real-time validation to reflect practical use cases in production-level applications. The "Company Title" field is mandatory, and leaving it empty should trigger an error message. Meanwhile, the "Number of Employees" field, though optional, should only accept numbers greater than zero, with errors displayed for invalid inputs. Additionally, the "Submit" button should remain disabled until the form is fully validated, aligning with best practices for guiding users toward correct form submissions.
\begin{figure}[H]
    \centering
    \includegraphics[width=0.8\textwidth]{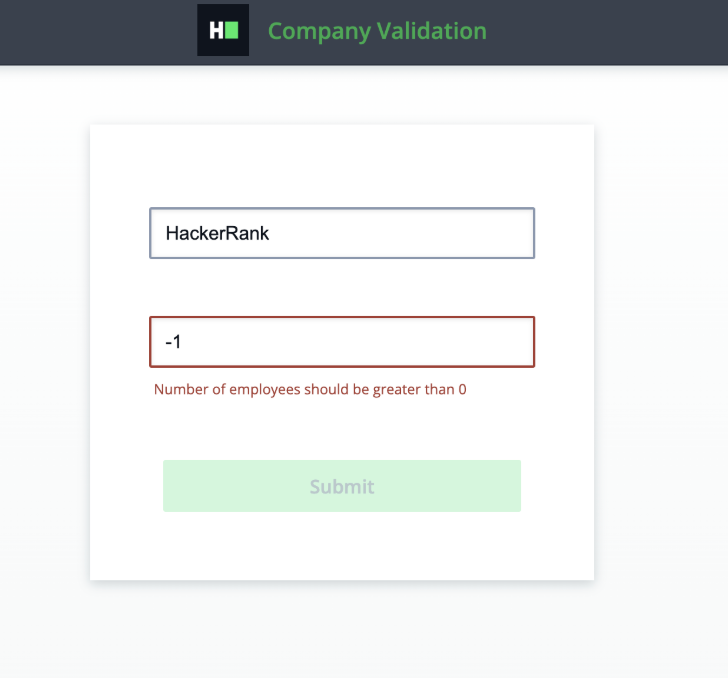} 
    \caption{A form is expected to behave dynamically by providing real-time feedback to users through form validation.}
\end{figure}
This scenario mirrors real-world front-end challenges in web development, where forms serve as a critical touchpoint for user interaction. Dynamic validation, error handling, and state management are essential for improving user experience and ensuring data accuracy. Proper implementation demands meticulous validation logic, efficient state handling, and conditional rendering of error messages.

A solution generated by GPT-4o demonstrated core competencies in state management and event handling but failed to execute dynamic validation and error management properly. The main shortcomings included the inability to dynamically re-evaluate the form’s validity, leading to incorrect enabling/disabling of the "Submit" button. Additionally, error messages were rendered even when empty, causing unnecessary DOM elements and test failures. By introducing a \texttt{useEffect} hook to monitor input and error state changes, conditionally rendering error elements, and updating error states dynamically during user input, these issues were rectified as the following screenshots. The corrected implementation adheres to production-grade standards, ensuring accurate validation and a seamless user experience.
\begin{figure}[H]
    \centering
    \includegraphics[width=0.8\textwidth]{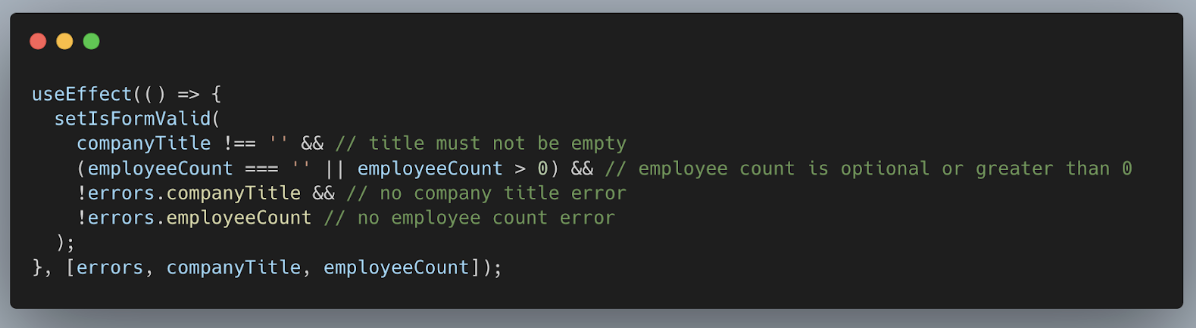} 
    \caption{A useEffect hook should be used to re-evaluate the form's validity whenever any error or input changes.}
\end{figure}
\begin{figure}[H]
    \centering
    \includegraphics[width=0.8\textwidth]{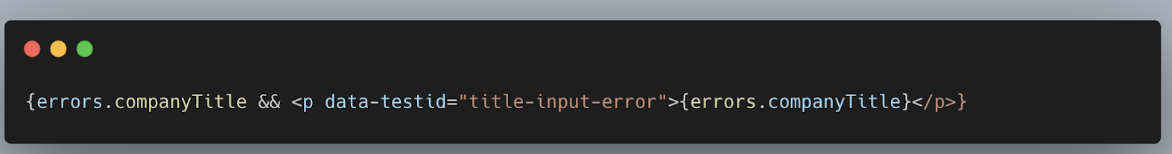} 
    \caption{Render error elements only when there is an actual error message.}
\end{figure}
\begin{figure}[H]
    \centering
    \includegraphics[width=0.8\textwidth]{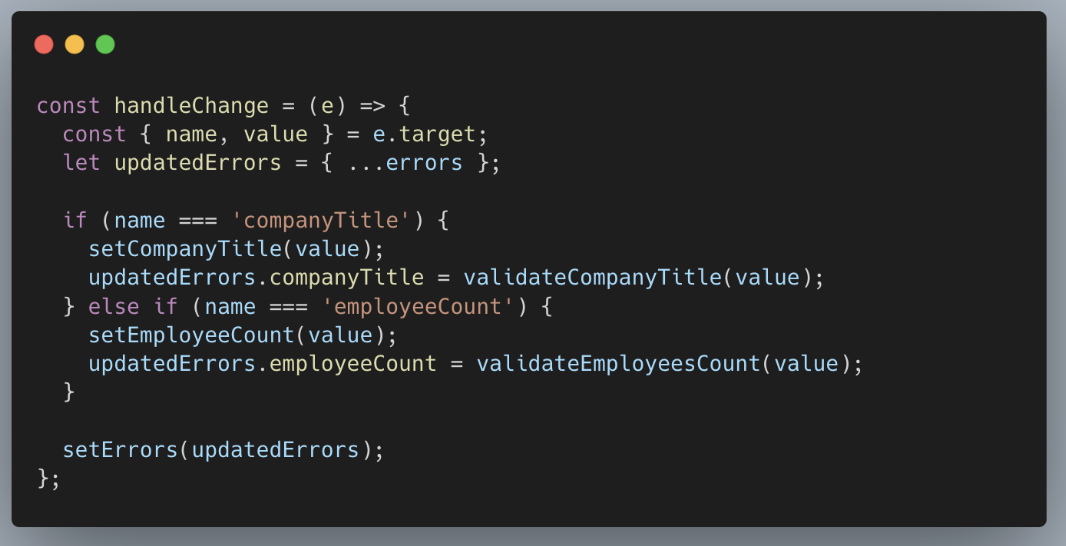} 
    \caption{The handleChange function should update the error state dynamically based on user input to ensure the UI always reflects the correct validation status.}
\end{figure}

\subsection{Errors Observed in a Angular project problem}
This task involves building a Length Converter component in Angular, where users can convert between units of length (e.g., Kilometers, Meters, and Centimeters) in real time. The component includes two input fields with corresponding dropdown menus for selecting units. When a user enters a value in one input field, the other field updates dynamically based on the selected units. Similarly, selecting a different unit from the dropdown menu recalculates the converted value for the associated input field. The initial state requires the first input field to default to "Kilometer" and the second to "Meter," with both input fields initially empty.
\begin{figure}[H]
    \centering
    \includegraphics[width=0.8\textwidth]{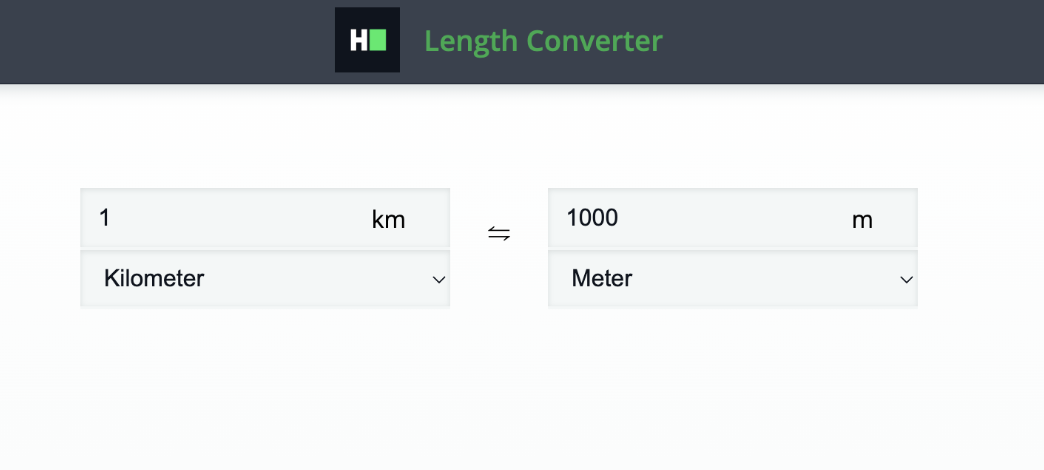} 
    \caption{Length Converter functionality.}
\end{figure}
This problem reflects real-world use cases for unit conversion, commonly seen in industries like e-commerce, architecture, and engineering, where accurate and dynamic conversions are critical. For example, online platforms often provide options to switch measurement units for materials, while navigation systems allow users to toggle between distance measurements such as kilometers and miles. This scenario requires developers to employ two-way data binding, lifecycle hooks, and conversion logic to synchronize inputs, dropdowns, and calculated results seamlessly.

\begin{figure}[H]
    \centering
    \includegraphics[width=0.8\textwidth]{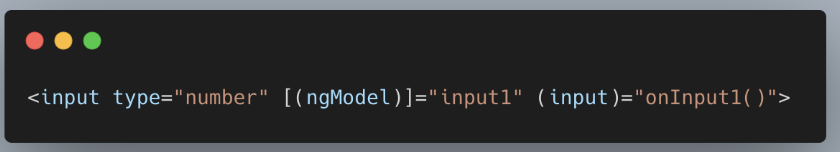} 
    \caption{This binds the input1 field to the component's input1 property. Any changes to the input field will trigger the onInput1() method for conversion.}
\end{figure}
\begin{figure}[H]
    \centering
    \includegraphics[width=0.8\textwidth]{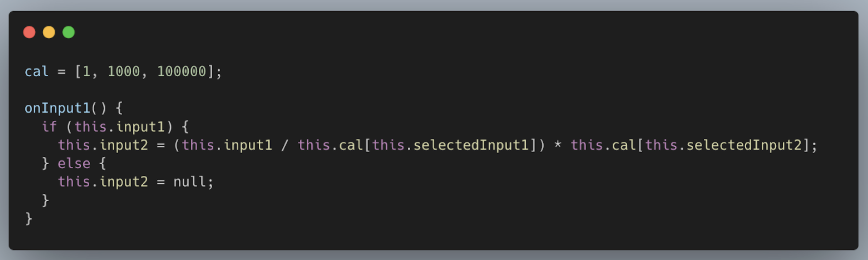} 
    \caption{The conversion logic should ensure that the corresponding input field is updated whenever a change occurs.}
\end{figure}
\begin{figure}[H]
    \centering
    \includegraphics[width=0.8\textwidth]{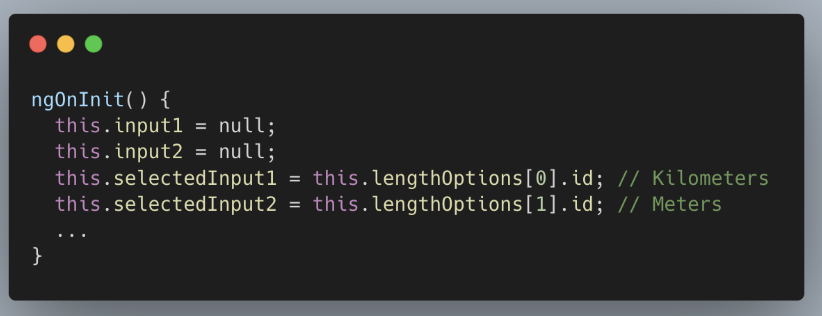} 
    \caption{The ngOnInit() lifecycle hook initializes the component’s state. Initially, the input1 and input2 should be set to null, with the default units being Kilometer and Meter.}
\end{figure}

A solution generated by GPT-4o successfully utilized Angular's two-way data binding with \texttt{ngModel} to link the input fields and dropdowns to component properties. It also implemented a conversion matrix to calculate factors between units. However, specific flaws impacted the functionality. The input fields were incorrectly initialized to 0 instead of being empty (\texttt{null}), violating the problem requirements. Additionally, dropdown changes occasionally updated the wrong input field, leading to unpredictable results. Null and undefined values were not adequately handled, causing errors during conversions. By addressing these issues—ensuring input fields are initialized to \texttt{null}, updating the correct field dynamically based on unit changes, and adding null checks—the component can function as intended, offering a seamless and accurate user experience.

\subsection{Errors Observed in a Ruby on Rails project problem}
This task involved developing a Job Board API using Ruby on Rails, allowing job seekers and employers to interact seamlessly. The API required implementing key functionalities, such as user authentication, job listings, job search, and job application submissions. Users could register and log in securely, with JWT tokens ensuring authenticated interactions. The job listings included advanced search capabilities using the Ransack gem, filtering results dynamically by title, company, or location. Additionally, authenticated users could apply for jobs by submitting CVs and cover letters, with file upload handling as part of the requirements.

The problem mirrors real-world web development tasks, such as securing APIs with JWT tokens, implementing RESTful design principles, and managing file uploads. These features are common in systems requiring secure access and efficient query tools, such as e-commerce or enterprise applications. Handling user registration, authentication, and advanced search showcases essential skills in API design, while ensuring data validation and response consistency is critical for production-grade systems.

A solution generated by GPT-4o successfully implemented core functionalities like user registration, login, job listings retrieval, and search. It efficiently utilized Ransack for advanced query filtering and followed RESTful architecture principles for modular and scalable endpoint design. However, key gaps impacted edge case handling and critical validations. For instance, the \texttt{current\_user} method was missing, preventing the application from identifying authenticated users in job application actions. Similarly, the \texttt{Job} model lacked validations for required fields, allowing invalid entries to be created. Authentication concerns arose when unauthenticated users could access protected routes due to missing the \texttt{AuthenticateUser} concern in the \texttt{JobsController}. Addressing these issues—such as adding the \texttt{current\_user} method, including field validations, and securing protected routes—resulted in a fully functional and secure application that passed all test cases.
\begin{figure}[H]
    \centering
    \includegraphics[width=0.8\textwidth]{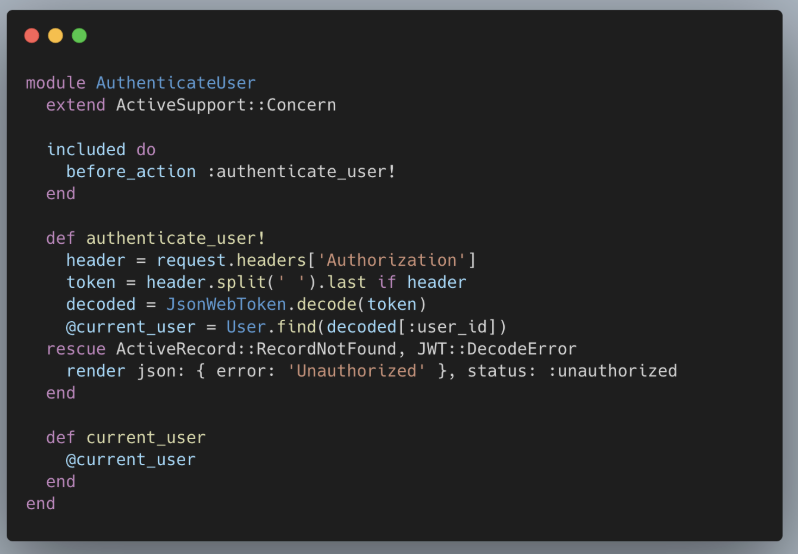} 
    \caption{Fix: Adding the current\_user method to the AuthenticateUser concern to ensure that the method is available for use in the controller actions.}
\end{figure}
\begin{figure}[H]
    \centering
    \includegraphics[width=0.8\textwidth]{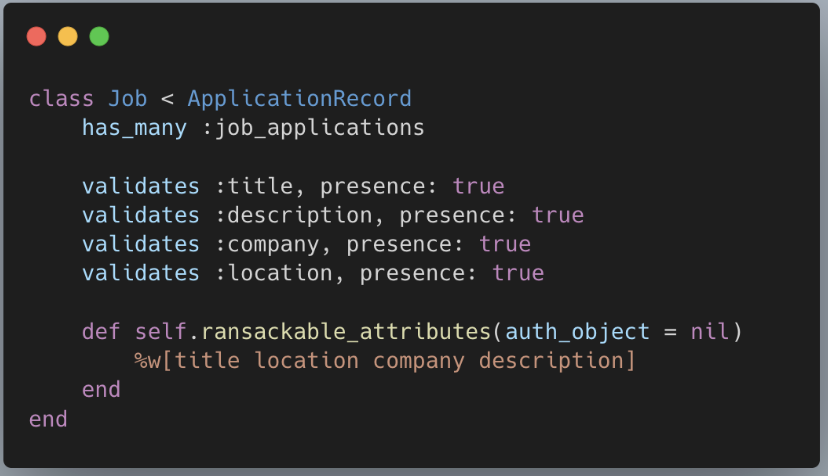} 
    \caption{Fix: Adding validations ensured that only valid job entries were created, improving data integrity.}
\end{figure}
\begin{figure}[H]
    \centering
    \includegraphics[width=0.8\textwidth]{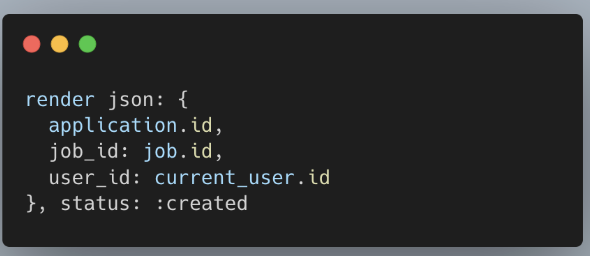} 
    \caption{Fix: The response was updated to include details about the associated job and user, making it more comprehensive.}
\end{figure}

\end{document}